\documentclass[12pt]{article}
\usepackage[utf8]{inputenc}
\usepackage{amsmath}
\usepackage{graphicx}
\usepackage{hyperref}
\usepackage{geometry}
\usepackage{setspace}
\usepackage{natbib}
\usepackage{titlesec}
\usepackage{tabularx}
\usepackage{adjustbox}
\usepackage{float}

\DeclareUnicodeCharacter{2212}{\textendash}

\titleformat{\section}{\large\bfseries}{\thesection.}{0.5em}{}
\titleformat{\subsection}{\normalsize\bfseries}{\thesubsection.}{0.5em}{}

\newcommand{\nopagenumbers}{%
  \clearpage
  \thispagestyle{empty}
  \setcounter{page}{0}
}

\begin{document}

% Title Page
\begin{titlepage}
    \begin{center}
        \vspace*{1in}
        {\Large \textbf{Buzz to Broadcast: Predicting Sports Viewership Using Social Media Engagement}}\\[1.5em]
        {\large Anakin Trotter}\\[1em]
        \textit{Department of Computer Science}\\
        \textit{The University of Texas at Austin}\\[2em]
        December 2, 2024\\[2em]
    \end{center}
\end{titlepage}
\nopagenumbers

% Abstract Section
\section*{Abstract}
Accurately predicting sports viewership is crucial for optimizing ad sales and revenue forecasting. Social media platforms, such as Reddit, provide a wealth of user-generated content that reflects audience engagement and interest. In this study, we propose a regression-based approach to predict sports viewership using social media metrics, including post counts, comments, scores, and sentiment analysis from TextBlob and VADER. Through iterative improvements, such as focusing on major sports subreddits, incorporating categorical features, and handling outliers by sport, the model achieved an \( R^2 \) of 0.99, a Mean Absolute Error (MAE) of 1.27 million viewers, and a Root Mean Squared Error (RMSE) of 2.33 million viewers on the full dataset. These results demonstrate the model's ability to accurately capture patterns in audience behavior, offering significant potential for pre-event revenue forecasting and targeted advertising strategies.

% Page Numbering Starts
\setcounter{page}{1}

\section{Introduction}

Understanding sports viewership is essential for guiding strategic and sales decisions in the sports broadcasting industry. It allows broadcasters to allocate resources efficiently, maximize revenue opportunities, and customize programming to align with viewer preferences. Regression, with its simplicity and reliability, has proven to be an effective tool for modeling audience ratings and identifying trends \citep{mahimkar2023enhancing}.

Social media platforms, such as Reddit, have emerged as valuable sources of user-generated content that reflect audience engagement and interest. These platforms offer a dynamic and real-time view of public opinion, providing an opportunity to explore how social media discussions can correlate with sports viewership. By analyzing metrics like post counts, comments, and sentiment, this study aims to uncover patterns that traditional methods may overlook.

In this research, we propose a novel regression-based approach that leverages social media engagement metrics to predict sports viewership. Sentiment analysis plays a key role, as it allows us to quantify public sentiment toward upcoming events. By combining sentiment features derived from TextBlob and VADER with advanced machine learning techniques, including gradient boosting regression and hyperparameter tuning, we aim to capture and model audience behavior with high accuracy.

Our findings demonstrate a strong correlation between Reddit activity and sports viewership, with higher engagement and positive sentiment corresponding to larger audiences. This research offers significant implications for pre-event revenue forecasting and targeted advertising strategies, providing broadcasters and advertisers with actionable insights to optimize their operations.

\section{Background}

Predicting sports viewership has long been a topic of interest in the private sector. For example, PredictHQ’s Live TV Events feature, led by Chief Data Officer Dr. Xuxu Wang, integrates multiple data sources, including sports team entities, county population data, broadcast schedules, league interest, and historical ratings. These features are combined within a probabilistic inference framework to predict viewership at the county level, providing businesses with actionable insights for inventory management, staffing, and advertising \citep{predicthq2022}. The model’s localized approach enables a granular understanding of audience behavior, showcasing the power of combining structured and unstructured data.

Similarly, Infinitive has explored AI-driven solutions to predict NFL viewership dynamics. Their approach incorporates variables such as Vegas odds, fantasy football statistics, and team win-loss records into advanced machine learning models. By accounting for the multifaceted nature of live sports events, Infinitive’s models enhance the accuracy of ad inventory forecasts for media companies, optimizing revenue and operational efficiency \citep{infinitive2023}. This initiative demonstrates how diverse data sources and AI techniques can improve the precision of viewership predictions, particularly in the context of live sports streaming.

These examples highlight the growing sophistication of viewership prediction methods, demonstrating the feasibility of identifying broader trends that influence sports viewership, such as localized audience interest and game-specific factors. However, neither approach incorporates data from social media posts, comments, or sentiment, which offer real-time insights into audience engagement and public opinion. Studying the value of these metrics could further enhance the accuracy and breadth of viewership prediction models in the future.

The rise of social media platforms has introduced a wealth of user-generated content that can serve as a real-time proxy for audience sentiment and engagement. Platforms like Reddit provide rich data sources in the form of posts, comments, and discussion threads, which are often centered around upcoming events. These interactions offer unique insights into audience behavior, which can complement traditional metrics. Leveraging this data requires robust natural language processing (NLP) tools capable of extracting meaningful features from text.

Sentiment analysis, a key component of this study, involves quantifying the polarity and subjectivity of text data. TextBlob, a Python-based library, employs a lexicon-based approach, utilizing pre-defined word lists and grammatical rules to compute sentiment scores. It assigns a polarity score ranging from -1 (negative) to +1 (positive) and a subjectivity score from 0 (objective) to 1 (subjective). TextBlob is particularly effective for general-purpose text analysis and has been widely adopted due to its simplicity and versatility \citep{loria2018textblob}.

VADER (Valence Aware Dictionary and sEntiment Reasoner), on the other hand, is a lexicon and rule-based sentiment analysis tool specifically designed for social media text. VADER incorporates intensifiers, such as capitalization and exclamation marks, and accounts for the nuances of informal language, emojis, and slang often present in social media posts. It outputs sentiment scores for positive, negative, and neutral sentiments, along with a compound score that reflects the overall sentiment polarity of the text \citep{hutto2014vader}. By combining TextBlob and VADER, this study aims to improve sentiment recall, as the two models sometimes yield different results due to their distinct methodologies.

The predictive modeling component of this research employs gradient boosting regression, an ensemble machine learning technique that builds models sequentially, with each iteration correcting the errors of its predecessor. Gradient boosting is particularly effective for handling complex, non-linear relationships in data, as demonstrated by \citet{friedman2001greedy}. The algorithm minimizes a specified loss function by iteratively adding weak learners, such as decision trees, to optimize predictions. This study uses the implementation provided by scikit-learn, which allows for flexibility in customizing model parameters \citep{pedregosa2011scikit}.

To optimize the performance of the gradient boosting model, hyperparameter tuning is conducted using GridSearchCV. This method systematically evaluates combinations of parameters, such as the number of estimators, learning rate, and maximum tree depth, to identify the configuration that minimizes prediction error. Hyperparameter tuning has been shown to significantly improve the generalization ability of machine learning models, particularly when dealing with complex datasets \citep{bergstra2012random}.

To interpret the contributions of features in the predictive model, we employ SHAP (SHapley Additive exPlanations), a method developed by \citet{lundberg2017unified}. SHAP is based on cooperative game theory and assigns each feature an importance value by calculating its marginal contribution to the model's predictions. Specifically, SHAP generates a set of additive feature attributions for each instance, ensuring consistency and local accuracy. This interpretability framework is particularly useful for complex models like gradient boosting, where the relationships between features and outcomes are non-linear and difficult to intuit. In this study, SHAP analysis highlights the significant predictive power of metrics such as total post count, comment count, and sentiment scores from both TextBlob and VADER, which play complementary roles in predicting viewership.

By integrating social media engagement metrics with advanced machine learning techniques, this research builds upon and extends existing work in viewership prediction. While prior studies have successfully modeled viewership using traditional metrics, this study explores how modern tools, such as sentiment analysis and gradient boosting regression, can provide new insights into audience behavior.

\section{Data Collection and Preparation}

This section outlines the sources and preprocessing steps used for TV viewership and Reddit activity data. Reddit was chosen due to having freely available historical data. These datasets were combined to study patterns and predict sports viewership trends.

\subsection{TV Viewership Data}

Viewership data were scraped from Wikipedia pages documenting television ratings for major sports events, including the World Series, Super Bowl, NBA Finals, Stanley Cup Finals, and MLS Cup. These pages were manually validated against TV viewership reports from \textit{SportsMediaWatch} (\url{https://www.sportsmediawatch.com/}) and \textit{Nielsen Media Research} (\url{https://www.nielsen.com/}). High-profile games were selected for their widespread appeal and availability of freely accessible data. For example, events like the Super Bowl or World Series attract millions of viewers, making publicly available ratings data reliable for initial modeling. Additionally, the large audiences reduce the impact of prediction errors since deviations of a few million viewers are relatively minor in the context of tens of millions.

While these public datasets provided a solid starting point, the availability of more granular or proprietary viewership data, such as detailed ratings from media corporations, could enable significant improvements in model accuracy and applicability. Table~\ref{tab:tv_viewership} provides an abbreviated summary of the TV viewership data. The full dataset is available in the appendix.

\begin{table}[htbp]
\centering
\caption{Abbreviated TV Viewership Data}
\label{tab:tv_viewership}
\begin{tabularx}{\textwidth}{|l|l|l|X|}
\hline
\textbf{Name} & \textbf{Teams} & \textbf{Start Time} & \textbf{Avg Viewers (M)} \\
\hline
WS G1 2024 & LAD, NYY & 10/25/24 5:11 PM & 14.16 \\
WS G2 2024 & LAD, NYY & 10/26/24 5:15 PM & 13.71 \\
WS G3 2024 & LAD, NYY & 10/28/24 5:17 PM & 13.21 \\
WS G4 2024 & LAD, NYY & 10/29/24 5:08 PM & 16.28 \\
WS G5 2024 & LAD, NYY & 10/30/24 5:08 PM & 18.15 \\
\hline
\end{tabularx}
\end{table}

\subsection{Reddit Activity Data}

To complement viewership data, Reddit activity surrounding these events was collected using the \textit{PullPush API} (\url{https://pullpush.io/}). Queries included the event title (e.g., ``World Series'' or ``Super Bowl'') and participating team names, restricted to sport-specific subreddits for improved relevance. Broad titles were used instead of specific game numbers (e.g., ``Super Bowl XLVI'') to capture a wider range of posts, as users in colloquial discussions are less likely to reference game numbers explicitly. For example, the query for the 2024 World Series would be:

\begin{quote}
``World Series'' OR ``LAD'' OR ``NYY''
\end{quote}

This strategy ensures that discussions about the overall event are included, capturing the general hype and sentiment rather than limiting results to highly specific terms. The use of a narrow time window (72 hours before the game) helps prevent previous or unrelated games from overly influencing the data, maintaining focus on the current game's context.

The challenges of social media analytics are particularly relevant to this paper due to the vast volume of data and the dynamic, semi-structured nature of user-generated content. Identifying posts relevant to specific sporting events is complicated by the high noise levels and lack of context in platform-wide searches. By narrowing the focus to specific subreddits, our research leverages the contextually organized and topic-driven nature of subreddit discussions, which provide a higher signal-to-noise ratio. This structured approach minimizes irrelevant data, ensures that the content is directly related to the events being analyzed, and capitalizes on community-driven curation to enhance the quality of the dataset. These methodological choices improve the efficiency and accuracy of data collection, aligning the social media data more closely with the viewership metrics \citep{stieglitz2018social}.

The PullPush API imposes a limit of 100 results per query. To handle this, a rolling time-window strategy was employed, capturing posts from the 72-hour period preceding each event. This window balanced relevance and coverage, though further studies could explore optimizing the time frame. For instance, alternative windows of 24 and 168 hours were tested but failed to produce viable models. A systematic approach to determining the ideal time window could enhance model performance in future research.

Reddit data were aggregated to include the total number of posts, comments, and scores (upvotes and downvotes) for each event. Sentiment analysis was performed using TextBlob and Vader to compute average sentiment scores. The dataset was filtered to exclude games with zero posts, though no such instances were observed in the data collection. Table~\ref{tab:reddit_activity} provides an abbreviated summary of the Reddit activity data, with the full dataset available in the appendix.

\begin{table}[htbp]
\centering
\caption{Abbreviated Reddit Activity Data}
\label{tab:reddit_activity}
\scriptsize
\begin{tabular}{|l|l|p{1.2cm}@{\hskip 0pt}|p{1.2cm}@{\hskip 0pt}|p{1.8cm}@{\hskip 0pt}|p{1cm}|p{1.3cm}|p{1.3cm}|p{1.5cm}@{\hskip 0pt}|}
\hline
\textbf{Name} & \textbf{Year} & \textbf{Teams} & \textbf{Posts} & \textbf{Comments} & \textbf{Scores} & \textbf{TextBlob Sent.} & \textbf{Vader Sent.} & \textbf{Viewers (M)} \\
\hline
SB XLVI & 2012 & NYG\newline NEP & 98 & 2067 & 1277 & 0.27 & 0.65 & 111.35 \\
SB XLVII & 2013 & BAL\newline SF & 107 & 2099 & 4923 & 0.27 & 0.67 & 108.69 \\
SB XLVIII & 2014 & SEA\newline DEN & 160 & 2237 & 3077 & 0.28 & 0.65 & 112.19 \\
\hline
\end{tabular}
\end{table}

\section{Feature Extraction}
\label{sec:feature_extraction}

As alluded to in the previous tables summarizing the TV viewership and Reddit activity data, feature extraction was a critical step in transforming raw data into structured inputs suitable for predictive modeling. This process involved identifying and engineering features that best capture audience engagement, sentiment, and sport-specific factors. The goal was to extract meaningful, non-redundant features that correlate strongly with viewership metrics while avoiding overfitting or multicollinearity.

The final set of features used in the regression model is as follows:
\begin{itemize}
    \item \textbf{Total Posts:} The number of posts made in the relevant subreddit within the 72-hour window before the event.
    \item \textbf{Total Comments:} The total number of comments on posts in the same time window.
    \item \textbf{Total Scores:} The aggregate score (upvotes minus downvotes) of posts during the time window.
    \item \textbf{Average Sentiment (TextBlob):} The average polarity score of all posts, as determined by TextBlob.
    \item \textbf{Average Sentiment (VADER):} The average compound sentiment score of all posts, as determined by VADER.
    \item \textbf{Sport:} A categorical variable indicating the type of game (e.g., World\_Series), encoded using one-hot encoding.
\end{itemize}

These features were chosen based on their hypothesized relationship with viewership. Subreddit activity metrics reflect overall engagement levels, sentiment scores capture the tone of discussions, and sport type addresses known variations in audience size across leagues.

\subsection{Numerical and Categorical Features}

The features were divided into two categories: numerical features that quantify engagement and sentiment, and categorical features representing the type of sport.

\begin{figure}[H]
\centering
\includegraphics[width=0.8\textwidth]{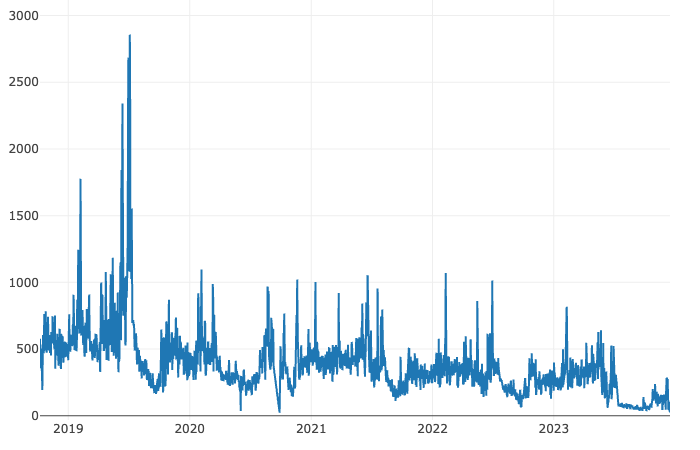}
\caption{Daily posting activity in r/nba, as obtained from Subreddit Stats. The data show stable engagement levels with an outlier spike in June 2019.}
\label{fig:nba_posts_per_day}
\end{figure}

Numerical features included \textit{Total Posts}, \textit{Total Comments}, \textit{Total Scores}, \textit{Average Sentiment (TextBlob)}, and \textit{Average Sentiment (VADER)}. These engagement metrics were left in their raw form based on evidence of stable subreddit activity over time. Data from \textit{Subreddit Stats} (\url{https://subredditstats.com/}) confirmed consistent posting activity for subreddits such as r/nba, with isolated spikes. Figure~\ref{fig:nba_posts_per_day} shows the daily posting activity in r/nba, highlighting an outlier spike in June 2019. Retaining raw values ensured that the inherent differences in subreddit engagement levels were preserved.

Sentiment scores were averaged across posts to reduce noise introduced by outliers and highly polarized individual posts. Averaging ensures that the feature reflects the overall tone of subreddit discussions, aligning with best practices in sentiment analysis \citep{feldman2013techniques}. Highly negative or positive posts can skew the results if treated individually, whereas averaging provides a more balanced and representative measure of community sentiment.

Categorical features, such as \textit{Sport}, were encoded using one-hot encoding. This method converts each category into a binary vector, allowing the model to treat categories independently without imposing artificial ordinal relationships (e.g., NBA\_Finals $>$ Stanley\_Cup). One-hot encoding is widely used in machine learning to ensure that non-ordinal data is represented effectively \citep{potdar2017categorical}.

\subsection{Multicollinearity Assessment}

\begin{figure}[H]
\centering
\includegraphics[width=0.8\textwidth]{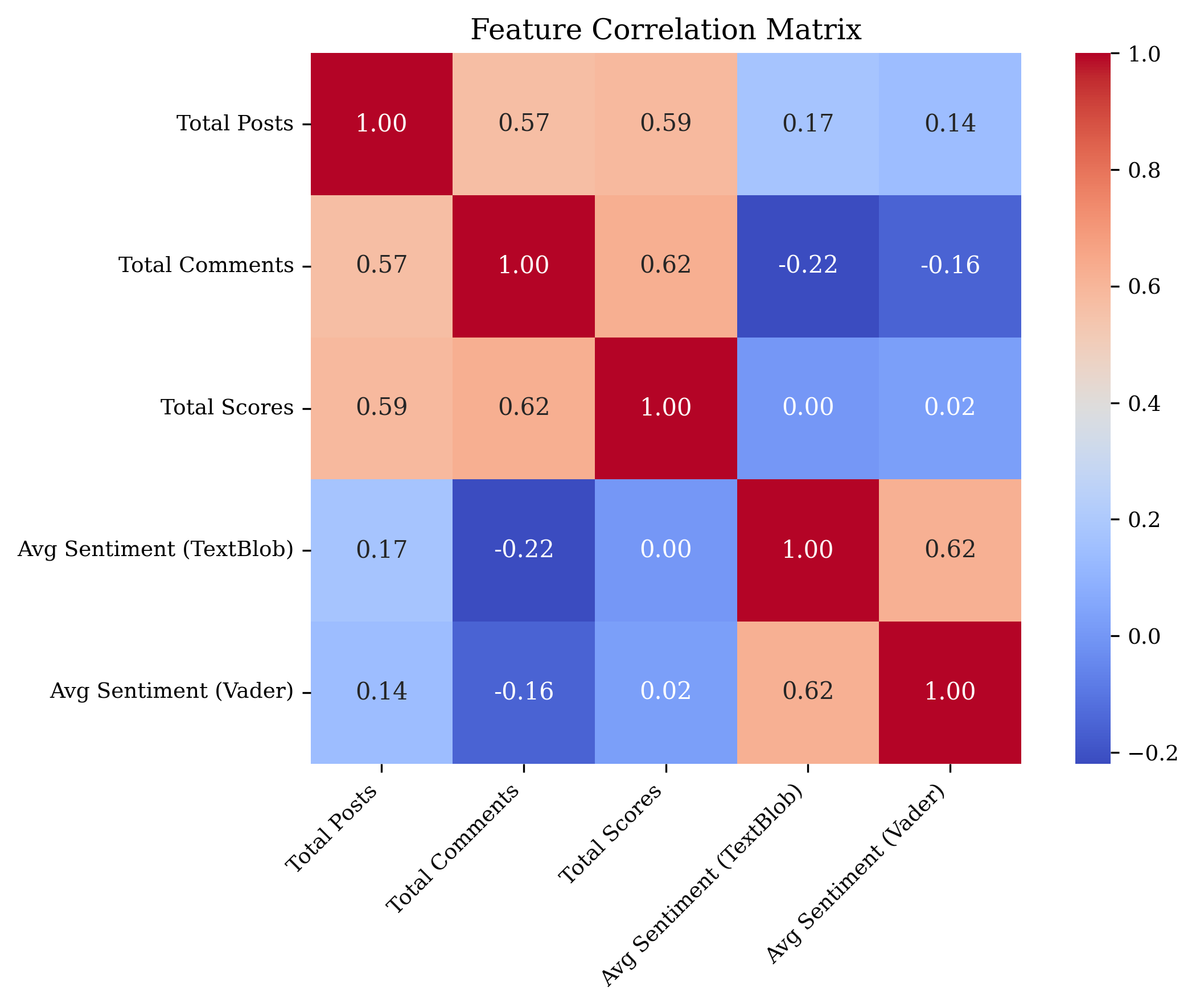}
\caption{Feature correlation matrix. The highest correlation observed was 0.62, indicating no strong multicollinearity.}
\label{fig:feature_correlation_matrix}
\end{figure}

Multicollinearity can undermine the reliability of regression models by inflating standard errors, biasing coefficients, and increasing the likelihood of Type 1 errors (false positives) \citep{kalnins2022multicollinearity}. To address this, a feature correlation matrix was generated, as shown in Figure~\ref{fig:feature_correlation_matrix}. The matrix revealed that no pair of features exceeded a Pearson correlation coefficient of 0.62, which is below the commonly accepted threshold of 0.7 for strong correlation \citep{ratner2009correlation}. This result indicates that the selected features are sufficiently independent to be used in the regression model without introducing significant multicollinearity.

Although multicollinearity is not as important when using gradient boosting models due to their ability to handle redundant features effectively, it can still impact interpretability and computational efficiency, especially with highly correlated variables. In the presence of multicollinearity, interpreting feature importance in gradient boosting models can be challenging, as correlated features may distribute importance among themselves, obscuring their true individual impact.

\subsection{Feature Relevance and Challenges}

Each feature was carefully chosen based on its potential to predict viewership. Subreddit engagement metrics, such as \textit{Total Posts} and \textit{Total Comments}, directly measure activity levels and are hypothesized to correlate with greater public interest and larger audiences. Sentiment scores add qualitative insight by capturing the tone of discussions, with positive sentiment often reflecting anticipation or excitement, and negative sentiment signaling controversy or dissatisfaction \citep{feldman2013techniques}. The inclusion of \textit{Sport} accounts for league-specific differences in viewership patterns, influenced by factors such as fan loyalty and game stakes \citep{mahimkar2023enhancing}.

Noise in sentiment data and outliers in subreddit activity posed challenges during feature engineering. Averaging sentiment scores mitigated the impact of extreme posts, while empirical assessments of subreddit activity stability guided the decision to retain raw engagement metrics. Future research could explore alternative feature transformations, such as log transformations for skewed distributions or interaction terms to capture non-linear relationships.

By combining numerical engagement metrics, sentiment analysis, and categorical sport data, this study builds a comprehensive feature set that effectively captures the factors influencing sports viewership. Careful preprocessing ensured that the input data were both representative and suitable for regression modeling.

\section{Methods}

This section details the methodology employed in training and evaluating the predictive model. The process involved selecting an appropriate regression algorithm, preprocessing data, tuning hyperparameters, and validating the model using robust metrics.

\subsection{Model Selection}

Gradient Boosting Regression (GBR) was chosen as the primary algorithm due to its ability to handle complex, non-linear relationships and its robustness against overfitting when properly regularized \citep{friedman2001greedy}. GBR builds models iteratively, optimizing a specified loss function by combining weak learners, typically decision trees. This technique is particularly effective for datasets with mixed feature types, such as numerical engagement metrics and categorical sports data.

The algorithm also provides interpretable outputs, including feature importance scores, which make it suitable for identifying the key drivers of sports viewership. GBR has been widely adopted in regression tasks involving predictive modeling \citep{natekin2013gradient}.

\subsection{Preprocessing and Train-Test Split}

The preprocessing steps prepared the features and target variable for training and evaluation, ensuring consistency across the pipeline. The following transformations were applied:

\begin{itemize}
    \item \textbf{Log Transformation:} The target variable, \textit{Viewers (Millions)}, was log-transformed using \( \log_{e}(1+x) \) to reduce the skewness in its distribution, a technique commonly used to normalize data for regression \citep{osborne2010improving}.
    \item \textbf{Outlier Removal:} Numerical features (\textit{Total Posts}, \textit{Total Comments}, \textit{Total Scores}, \textit{Average Sentiment (TextBlob)}, \textit{Average Sentiment (VADER)}) were screened for outliers using the interquartile range (IQR) method, as described in Section~\ref{sec:feature_extraction}. This process ensured that extreme values did not disproportionately influence the model.
    \item \textbf{Scaling:} Numerical features were scaled to the [0, 1] range using Min-Max Scaling, which preserves relative differences between values while standardizing feature magnitudes \citep{han2011data}.
    \item \textbf{One-Hot Encoding:} The categorical feature \textit{Sport} was one-hot encoded, creating binary columns for each sport (e.g., \textit{World\_Series}, \textit{Super\_Bowl}). This approach avoids imposing ordinal relationships, as discussed in Section~\ref{sec:feature_extraction} \citep{potdar2017categorical}.
\end{itemize}

After preprocessing, the dataset was split into training (80\%) and testing (20\%) sets using \texttt{train\_test\_split} from \texttt{scikit-learn}, with a random state of 42 for reproducibility \citep{pedregosa2011scikit}.

\subsection{Hyperparameter Tuning}

Hyperparameter tuning was conducted using \texttt{GridSearchCV} to systematically evaluate combinations of parameters and optimize the model’s performance. The tuning process focused on the following parameters for the Gradient Boosting Regressor:

\begin{itemize}
    \item \textbf{\texttt{n\_estimators}:} Number of boosting stages, evaluated at 100 and 200.
    \item \textbf{\texttt{learning\_rate}:} Shrinkage factor for updating weights, evaluated at 0.05.
    \item \textbf{\texttt{max\_depth}:} Maximum depth of individual trees, evaluated at 3 and 5.
    \item \textbf{\texttt{min\_samples\_split}:} Minimum samples required to split an internal node, evaluated at 2 and 5.
    \item \textbf{\texttt{subsample}:} Fraction of samples used for fitting individual base learners, evaluated at 0.8 and 1.0.
\end{itemize}

The grid search employed 5-fold cross-validation with \texttt{neg\_mean\_absolute\_error} as the scoring metric. Cross-validation ensures that the model generalizes well to unseen data, reducing the risk of overfitting \citep{hastie2009elements}. The best hyperparameters were selected based on performance on the training set and stored for subsequent evaluation.

\subsection{Evaluation Metrics}

The model’s performance was evaluated using three complementary metrics, applied to inverse log-transformed predictions and actual test data:

\begin{itemize}
    \item \textbf{Mean Absolute Error (MAE):} Measures the average absolute difference between predicted and actual values, providing an interpretable measure of error in millions of viewers \citep{willmott2005advantages}.
    \item \textbf{Root Mean Squared Error (RMSE):} Penalizes larger errors more heavily than MAE by squaring differences before averaging, offering insights into outlier sensitivity.
    \item \textbf{\( R^2 \) Score:} Represents the proportion of variance in the dependent variable explained by the model. This metric evaluates the overall goodness of fit \citep{hastie2009elements}.
\end{itemize}

These metrics collectively provide a robust assessment of the model’s accuracy and error distribution. MAE captures typical error magnitude, RMSE highlights the influence of larger deviations, and \( R^2 \) evaluates the overall explanatory power of the model.

\section{Results}

This section delves into the model's performance, the relative importance of features, and the implications of observed trends in the data.

\subsection{Model Performance}

The final model achieved the following performance metrics on the test set:
\begin{itemize}
    \item \textbf{Mean Absolute Error (MAE):} 1.27 million viewers
    \item \textbf{Root Mean Squared Error (RMSE):} 2.33 million viewers
    \item \textbf{\( R^2 \) Score:} 0.99
\end{itemize}

\begin{figure}[ht!]
\centering
\includegraphics[width=0.8\textwidth]{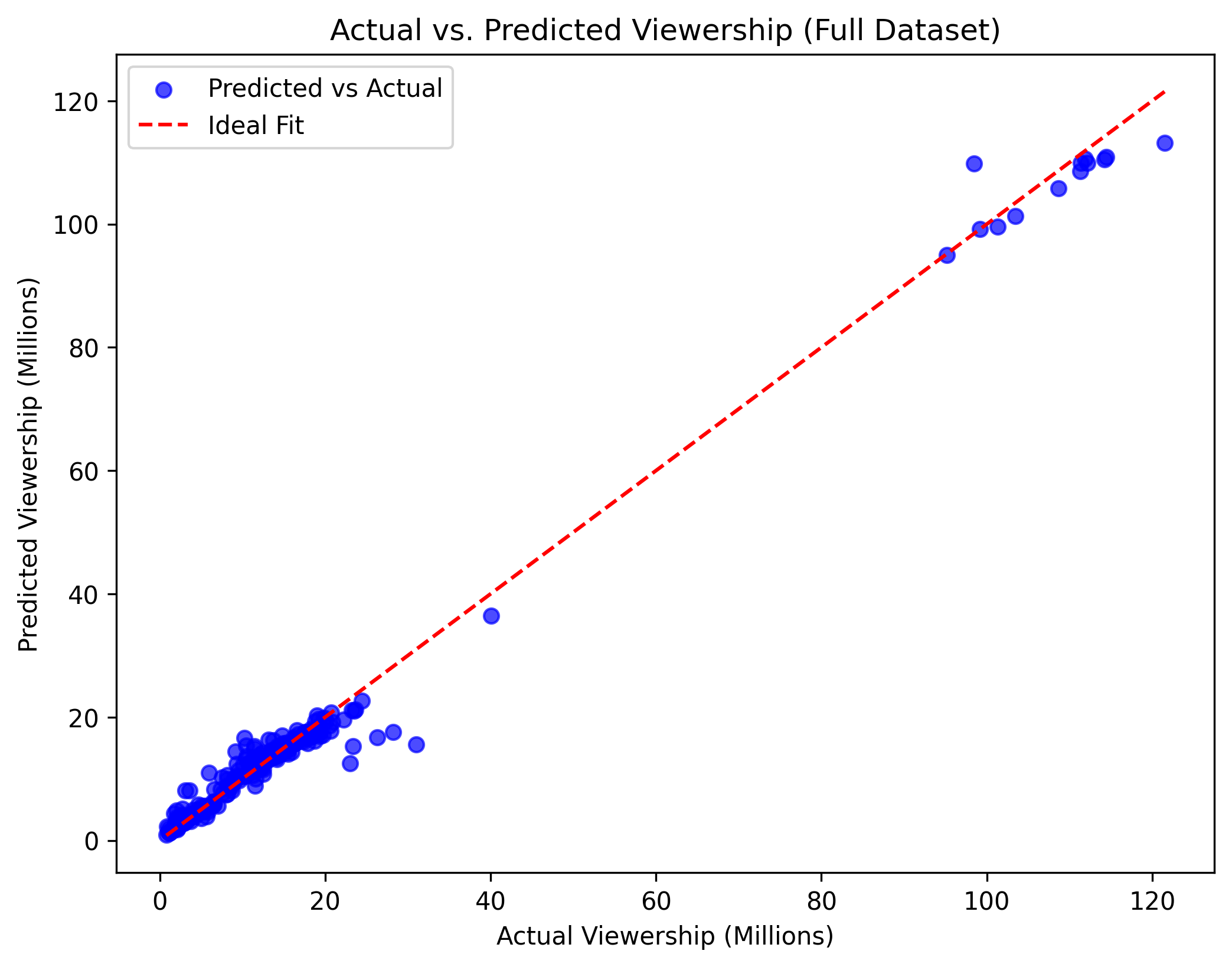}
\caption{Actual vs. Predicted Viewership. The red line represents an ideal fit, while blue points show model predictions. Stratification is visible between Super Bowl and other events.}
\label{fig:actual_vs_predicted}
\end{figure}

These results demonstrate that the model performs well, with high predictive accuracy and low error margins. The \( R^2 \) score of 0.99 indicates that 99\% of the variance in viewership is explained by the model. However, an evaluation of the actual vs. predicted values, visualized in Figure~\ref{fig:actual_vs_predicted}, reveals potential stratification in the data.

The stratification in Figure~\ref{fig:actual_vs_predicted} is most evident between Super Bowl games (with approximately 100 million viewers) and other events (less than 20 million viewers). While the model captures these patterns effectively, the disparity in viewership scales may influence the model's ability to generalize across sports. Future work could explore developing separate models for each sport to enhance prediction accuracy for events with fewer viewers.

\subsection{Feature Importance}

\begin{figure}[H]
\includegraphics[width=0.95\textwidth]{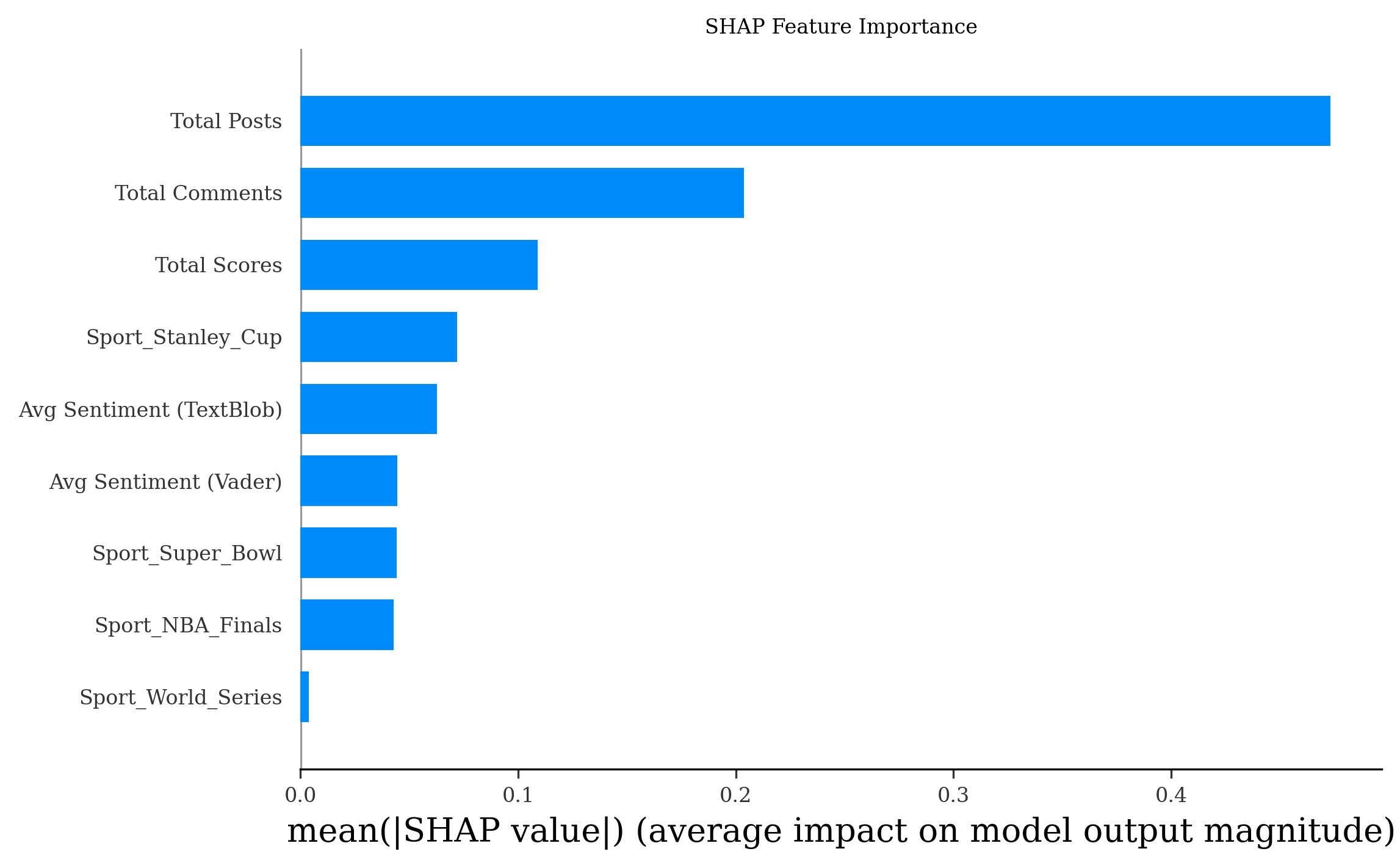}
\caption{SHAP Feature Importance Plot. The chart indicates that Total Posts is the most critical predictor, followed by Total Comments and Total Scores.}
\label{fig:shap_importance}
\end{figure}

Feature importance analysis, conducted using SHAP, highlights the relative contribution of each feature to the model's predictions. The SHAP summary plot in Figure~\ref{fig:shap_importance} shows that \textit{Total Posts} is the most influential feature by a significant margin, followed by \textit{Total Comments} and \textit{Total Scores}.

The dominance of social media metrics, particularly \textit{Total Posts}, underscores the strong correlation between audience engagement on Reddit and sports viewership. This result aligns with the study's hypothesis that social media activity is a robust predictor of audience interest. \textit{Total Comments} and \textit{Total Scores} further emphasize the value of user-generated content, with higher engagement metrics corresponding to larger audiences.

\section{Discussion}

The results highlight the potential of social media engagement metrics, particularly \textit{Total Posts}, as strong predictors of sports viewership. The model achieved high accuracy with an \( R^2 \) of 0.99 demonstrating its ability to explain the variance in viewership effectively. However, the stratified nature of the data, where Super Bowl games dominate the upper end of the viewership spectrum, presents challenges for generalizing predictions to smaller-scale events. The Mean Absolute Error (MAE) of 1.27 million viewers limits the model's accuracy for games with only a few million viewers, suggesting that it is better suited for predicting outcomes of highly popular games.

This study's reliance on Reddit data, while valuable for understanding engaged audiences, introduces a platform-specific bias and may not fully represent the broader public. Different demographics and user behaviors across platforms could mean that additional insights lie untapped in complementary data sources, such as Twitter or Instagram. Advanced sentiment analysis tools, such as context-aware models like BERT (Bidirectional Encoder Representations from Transformers), could improve predictive accuracy by capturing nuanced emotional expressions and context in audience discussions. These models can analyze the interplay between words, enabling deeper sentiment insights than traditional lexicon-based methods.

The small dataset of TV ratings, focused exclusively on high-profile games, also restricts the generalizability of these findings. Including smaller events and regular-season games could provide a more balanced view of audience behavior. Additionally, private companies with access to proprietary viewership data could achieve more tailored and accurate predictions. Such data would allow for league-specific or network-specific insights, while the publicly available data used in this study spanned multiple networks, limiting its granularity. For example, the ability to analyze viewership patterns within a single network could uncover programming-specific factors that influence audience size.

The use of a 72-hour time window for Reddit data collection further limits the findings, as this duration balances capturing relevant posts with excluding unrelated content. While effective for proof-of-concept modeling, future studies could explore different time windows or dynamic adjustments based on the nature of the event and its online engagement patterns.

Future research could address these limitations by building sport-specific models to better account for differences in audience behavior across leagues and events. Incorporating additional features, such as broadcast schedules, team popularity, and historical trends, could provide a more comprehensive understanding of viewership drivers. Expanding the analysis to include real-time social media data from multiple platforms could also enable dynamic forecasting during events, offering valuable insights for broadcasters and advertisers.

\section{Conclusion}

This study demonstrated the potential of social media engagement metrics, such as post counts, comments, and sentiment, in predicting sports viewership. While the model performed well for large-scale events, its limitations for smaller games highlight the importance of addressing dataset imbalances and incorporating more diverse data sources. By leveraging broader datasets, advanced sentiment analysis techniques, and multi-platform integration, future research can refine predictive models and enhance their applicability across different sports and audience scales.

\newpage

% Bibliography
\nocite{pullpush,sportsmediawatch,nielsen,mlb_wikipedia,nfl_wikipedia,nba_wikipedia,nhl_wikipedia,mls_wikipedia}
\nocite{subredditstats_nba,subredditstats_nfl,subredditstats_mlb,subredditstats_nhl,subredditstats_mls}
\bibliographystyle{apalike}
\bibliography{references}

\section*{Appendix}

\subsection*{Supplementary Materials}

The supplementary materials for this study include the TV viewership dataset and the code used for data preprocessing, modeling, and evaluation. These resources are publicly available and can be accessed through the following links:

\begin{itemize}
    \item \textbf{TV Viewership Data:} \href{https://docs.google.com/spreadsheets/d/1DLhuSqkbjLfA1q5lCfVsgoXgkWIUzIohc6GtT_TuB94/edit?usp=sharing}{Google Sheets - TV Viewership Data}
    \item \textbf{Code Repository:} \href{https://github.com/AnakinTrotter/reddit-sports-viewership-model}{GitHub - Reddit Sports Viewership Model}
\end{itemize}

These materials provide the raw data and implementation details necessary for replicating the study or conducting further analysis.

\end{document}